\documentclass{article} % For LaTeX2e
\usepackage{nips14submit_e,times}
\usepackage{hyperref}
\usepackage{url}
\usepackage{graphicx}
\usepackage[linesnumbered,ruled,vlined]{algorithm2e}
\usepackage{listings}
\usepackage{amsmath}
\usepackage{amssymb}

\usepackage{color}
\definecolor{dkgreen}{rgb}{0,0.6,0}
\definecolor{gray}{rgb}{0.5,0.5,0.5}
\definecolor{mauve}{rgb}{0.58,0,0.82}
 
 \lstdefinelanguage{scala}{
  morekeywords={abstract,case,catch,class,def,%
  do,else,extends,false,final,finally,%
  for,if,implicit,import,match,mixin,%
  new,null,object,override,package,%
  private,protected,requires,return,sealed,%
  super,this,throw,trait,true,try,%
  type,val,var,while,with,yield},
  otherkeywords={=>,<-,<\%,<:,>:,\#,@},
  sensitive=true,
  morecomment=[l]{//},
  morecomment=[n]{/*}{*/},
  morestring=[b]",
  morestring=[b]',
  morestring=[b]"""
}
 
\title{Factorbird - a Parameter Server Approach to Distributed Matrix Factorization}

\author{
Sebastian Schelter\thanks{work done while at Twitter} \\
Technische Universit{\"a}t Berlin\\
\texttt{@sscdotopen} \\
\And
Venu Satuluri \\
Twitter \\
\texttt{@vsatuluri} \\
\And
Reza Bosagh Zadeh\\
Stanford University\\
\texttt{@reza\_zadeh} \\
}

\nipsfinalcopy % Uncomment for camera-ready version

\begin{document}
\SetEndCharOfAlgoLine{}

\maketitle

\begin{abstract}
We present  `Factorbird', a prototype of a parameter server approach for factorizing large matrices with Stochastic Gradient Descent-based algorithms. 
We designed Factorbird to meet the following desiderata: (a) scalability to tall and wide matrices with dozens of
billions of non-zeros, (b) extensibility to different kinds of models and loss functions as long as they can be
optimized using Stochastic Gradient Descent (SGD), and (c) adaptability to both batch and
streaming scenarios. 
Factorbird uses a parameter server in order to scale to models that exceed the memory of an individual machine,
and employs lock-free Hogwild!-style learning with a special partitioning scheme to drastically reduce conflicting updates. 
We also discuss other aspects of the design of our system such as how to efficiently grid search for hyperparameters at scale. 
We present experiments of Factorbird on a matrix built from a subset of Twitter's interaction graph, 
consisting of more than 38 billion non-zeros and about 200 million rows and columns, which is to the best of our knowledge the
largest matrix on which factorization results have been reported in the literature.
\end{abstract}

\section{Introduction}

In recent years, there is a growing trend to apply machine learning (ML) to massive and complex datasets \cite{Amatriain2012}. For many applications, this leads to growing model sizes. A prime example are recommender systems where the size of the model is usually proportional to the number of users in the dataset \cite{Koren2009, Zhou2008}. In cases with hundreds of millions of users this model exceeds the memory of an individual machine and distributed approaches to model training that leverage multiple machines become necessary. This leads to a set of challenges, e.g. how to partition the model among the participating machines and how to correctly execute learning algorithms in such a setting.\\
In this work, we describe 'Factorbird', a prototypical system that leverages a parameter server
architecture~\cite{Li2014} for learning large matrix factorization models for recommendation mining. After a short
introduction to matrix factorization for recommender systems (Sec.~\ref{sec:latentfactormodels}), we describe the main
challenges of our system, namely how to partition the model among the machines in the cluster and how to run Stochastic
Gradient Descent (SGD) in parallel. We discuss the design decisions taken in Factorbird to overcome these challenges
(Sec.~\ref{sec:systemdesign}). First, we partition one of the matrices to learn over dedicated machines in the cluster
and co-partition the other one with the input data to localize a large number of updates and drastically reduce update
conflicts and network traffic. Next, we apply a lock-less Hogwild!-style~\cite{Recht2011} execution scheme to
efficiently run SGD in parallel. After giving insights into our software design and memory-efficient datastructures
(Sec.~\ref{sec:implementation}), we describe techniques to assess model quality in Factorbird
(Sec.~\ref{sec:modelquality}). Our approach here is to grid search over a large number of hyperparameter combinations in
a single training run. Finally, we present a set of experiments on user interaction data from twitter
(Sec.~\ref{sec:experiments}). We run a scale-out experiment on a matrix built from a subset of Twitter's interaction
graph~\cite{Kamath2014}, with more than 38 billion non-zeros and about 200 million rows and columns. To the best of our knowledge, this is largest dataset on which matrix factorization experiments have been published so far.

\section{Background: Latent Factor Models}
\label{sec:latentfactormodels}

\begin{figure}[h]
  \begin{center}
  \includegraphics[scale=0.33]{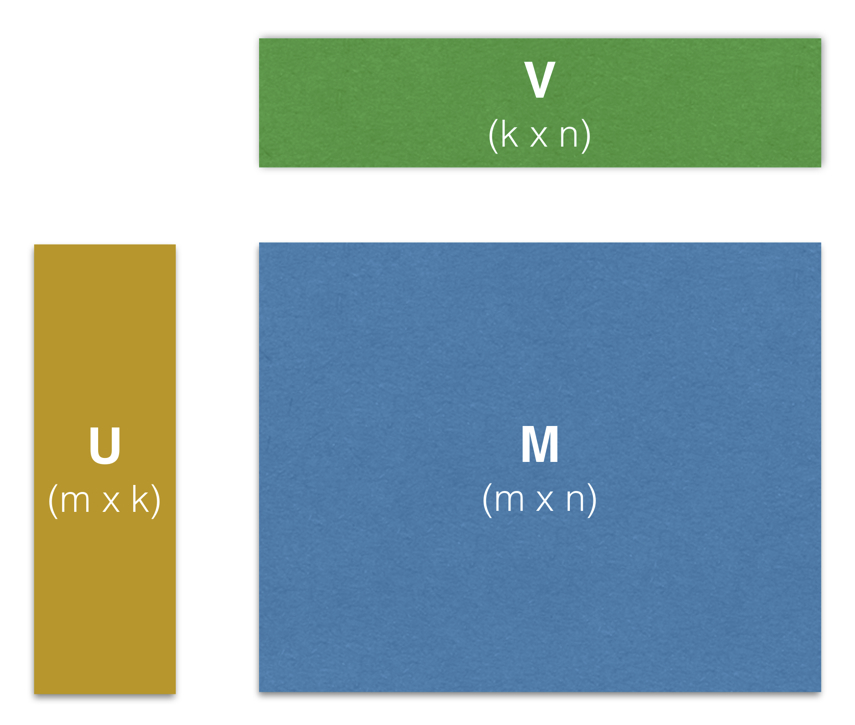}
  \end{center}
  \caption{Latent factor models as matrix factorization problem.}
  \label{fig:matrixfactorization}
\end{figure}

We build upon \textit{latent factor models}, which leverage a low-rank matrix factorization of interaction data \cite{Koren2009} to characterize users and items by vectors of factors inferred from interaction patterns. These methods have been among the top-scoring contributions to the Netflix prize \cite{Bennett2007}. They factor a sparse partially-observed $m \times n$ matrix $M$, representing the interactions of $m$ users with $n$ items, into the product of two rank $k$ factor matrices $U$ and $V$, such that their product $UV$ approximates the observed parts of $M$ and generalizes well to unobserved parts of $M$ (c.f. Figure~\ref{fig:matrixfactorization}).  A user $i$ is associated to a factor vector $u_i \in \mathbb{R}^k$  (a row of the $m \times k$ matrix $U$), while an item $j$ is associated to a factor vector $v_j \in \mathbb{R}^k$ (a column of the $n \times k$ matrix $V$). The factorization maps users and items onto joint latent factor space of low dimensionality $k$, such that  $u_i^T v_j$ estimates the strength of interaction between user $i$ and item $j$. 

The standard approach to learning a latent factor model for recommender systems is to minimize the regularized squared error over the predictions to the observed parts of the matrix \cite{Koren2009}:
$$\min\limits_{U,V} \sum\limits_{(i,j) \in M} (m_{ij} - u_i^T v_j)^2 + \lambda \left( \|u_i\|^2 + \|v_j\|^2 \right )$$
This approach is closely related to the Singular Value Decomposition (SVD) of the matrix, which gives an optimal rank $k$ approximation (w.r.t. to squared error) of the original matrix using the top-$k$ singular values and singular vectors. The key difference here is that the SVD is undefined when there are missing entries.\\
We adopt a more sophisticated approach outlined in \cite{Koren2009}. First, we introduce a global bias term $g$, which captures the average interaction strength between a user and an item in the dataset. Next, we introduce a user-specific bias term $b^U_i$ for user $i$ and an item-specific bias term bias $b^V_j$ for item $j$. These biases model how strongly the interactions of certain users and items tend to deviate from the global average. We substitute the dot product $u_i^T v_j$ for predicting the strength of the interaction between a user $i$ and an item $j$ with a prediction function $p(i,j) = g + b^U_i + b^V_j + u_i^T v_j$ that takes the bias terms into account. Furthermore, we introduce a function $a(i,j)$ for determining the strength of the interaction between user $i$ and item $j$. This allows us to transform the observed entries $m_{ij}$ of $M$. Finally, we add a function $w(i,j)$ for weighting the prediction error for interaction between user $i$ and item $j$. This function becomes useful when $M$ consists of data from various sources with differing confidence in the observations. The loss function that we minimize for our latent factor model in Factorbird is the following:
$$\min\limits_{g, b^{U}, b^{V}, U, V} \frac{1}{2} \left( \sum\limits_{i,j \in M} w(i,j) (p(i,j) - a(i,j))^2 \right) + \frac{\lambda}{2} \left( g^2 + 
  \|b^{U}\|^2 + \|b^{V}\|^2  + \|U\|^2_F + \|V\|^2_F  \right)$$
We adopt a graph-specific terminology for the rest of the paper as well as for the APIs of our system, as we think about the majority of Twitter's datasets in terms of large networks. Therefore, we assume that $M$ represents a network (e.g. the network of user followings in Twitter), $i$ and $j$ reference vertices in this network (e.g. two users in this network) and the edges of the network correspond to observed entries of $M$, meaning that $a(i,j)$ depicts the weight of a directed edge between a vertex $i$ and a vertex $j$ in this network. Furthermore, we assume that the bias vectors $b^U$ and $b^V$ are stored in $U$ and $V$ (e.g., as first column and first row), to simplify notation. 

\textbf{Factorization via Stochastic Gradient Descent (SGD).} There are various popular techniques to compute a matrix factorization. Stochastic Gradient Descent (SGD) \cite{Koren2009} randomly loops through all observed interactions, computes the error of the prediction for each interaction and modifies the model parameters in the opposite direction of the gradient. Another technique is Alternating Least Squares (ALS) \cite{Hu2008, Weimer2007, Zhou2008}, which repeatedly keeps one of the unknown matrices fixed, so that the other one can be optimally re-computed. We chose SGD as optimization technique, as it is simple, provides fast convergence and is easy to adapt to different models and loss functions. Algorithm \ref{alg:plainsgd} shows the individual steps to conduct when learning the matrix factorization with SGD. First, we randomly initialize the factor matrices $U$ and $V$ (c.f. line 1). Next, we randomly pick an edge $(i,j)$ from the graph and compute the weighted error $e_{ij}$ of the prediction $p(i,j)$ against the actual edge strength $a(i,j)$ (c.f. lines 3 \& 4). Next, we update the global bias term as well as the bias terms for $i$ and $j$ proportional to the prediction error $e_{ij}$, the learning rate $\eta$ and the regularization constant $\lambda$ (c.f. lines 5 to 7). We weight the regularization updates according the out-degree $n_i$ of vertex $i$ (the number of observed entries in the $i$-th row of $M$) and the  in-degree $n_j$ of $j$ (the number of observed entries in the $j$-th column of $M$). We update the factor vectors $u_i$ and $v_j$ analogously (c.f. lines 8 \& 9). The whole process is repeated until convergence.

\begin{algorithm}[H]
 randomly initialize $U$ and $V$\;
 \BlankLine\;
 \While{not converged}{
  randomly pick edge $(i,j)$\;
  \BlankLine\;
  \tcp{compute weighted prediction error}
  $e_{ij} \leftarrow w(i,j)  (a(i,j) - p(i,j))$ 
  \BlankLine\;
  \tcp{update biases}
 $g \leftarrow g - \eta \left(e_{ij} +\lambda g\right)$\;
 $b^U_i \leftarrow b^U_i - \eta \left(e_{ij} + \frac{\lambda}{n_i}b^U_i \right)$\; 
 $b^V_j \leftarrow b^V_j - \eta \left(e_{ij} + \frac{\lambda}{n_j}b^V_j \right)$\;
  \BlankLine\;
  \tcp{update factors}
 $u_i \leftarrow u_i - \eta \left(e_{ij} \; v_j + \frac{\lambda}{n_i}u_i \right)$\;
 $v_j \leftarrow v_j - \eta \left(e_{ij} \; u_i + \frac{\lambda}{n_j}v_j \right)$\;
 }
 \caption{Matrix Factorization using SGD.}
 \label{alg:plainsgd}
\end{algorithm}

\section{System Design}
\label{sec:systemdesign}

Having introduced the conceptual background of the models that we wish to learn, we proceed with describing the three main design goals of Factorbird.\\
First, Factorbird must handle factorizations of twitter-scale graphs with hundreds of millions of vertices and dozens of billions of edges. Scalability to datasets of this size is more important than high performance on small datasets commonly used in research (such as the Netflix \cite{Bennett2007} dataset). The matrices representing these graphs are either square (e.g. user to user followings) or `tall-and-wide' for bipartite graphs (e.g. user to tweets). Second, the system has to be highly usable and adaptable for data scientists without making a systems background a necessity. No systems programming should be required to try out a variation of our model or a different loss function. Instead, the system should offer interfaces that abstract from the distributed execution model and are intuitive to implement given an ML background. Third, the system design shall be simple to keep the maintenance and debugging effort low. Additionally, the system design should be easily extendable into a streaming system in the future, where a previously learned matrix factorization is updated online given a stream of new observations. This is the main reason why decided against using ALS, which is much harder to adapt to a streaming scenario than SGD.

\textbf{Challenges.} Running SGD on datasets of such scale inevitably leads to a set of challenges that have to be overcome when designing a scalable system. The main focus of this paper is to present the prototype of a system that elegantly solves these challenges.

(1) The resulting factor matrices for a huge network quickly become larger than the memory available on an individual commodity machine \cite{Hastie2014,Boyd2014}. For example, $U$ and $V$ with $k= 100$ and a single precision factor representation for a graph with 250 million vertices already have a combined size of about 200 GB. This estimation does not even take operating system buffers and caches, object references and required statistics like degree distribution of the graph into account, which also compete for memory.

(2) Due to the sheer number of observations in large datasets, we aim to leverage multiple cores and machines to learn our desired matrix factorization. Unfortunately, SGD is an inherently sequential algorithm. It randomly picks an observation and updates the model parameters before proceeding to the next observation (c.f. Algorithm \ref{alg:plainsgd}). When we run SGD in parallel on multiple cores, there is a chance that we concurrently try to update the same $u_i$ or $v_j$, which results in conflicting writes and lost updates. 

\begin{figure}[h]
  \begin{center}
  \includegraphics[scale=0.33]{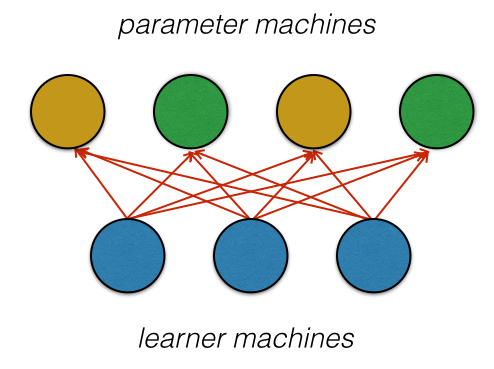}
  \end{center}
  \caption{Parameter Server architecture.}
  \label{fig:paramserver}
\end{figure}

In order to overcome the first challenge, we decided for a distributed architecture that allows us to partition the large model (the factor matrices) over several machines. We adapt a `Parameter Server' architecture \cite{Li2014}. As illustrated in Figure~\ref{fig:paramserver}, we partition the factor matrices over a set of machines, to which we refer to as \textit{parameter machines}. At the same time, we partition the graph (our input data) over a set of so-called \textit{learner machines}. Each learner machine runs multi-threaded SGD on its portions of the input data. For every observation to process, the learner machine has to fetch the corresponding factor vectors from the parameter machines, update them and write them back over the network.

\begin{figure}[h]
  \begin{center}
  \includegraphics[scale=0.4]{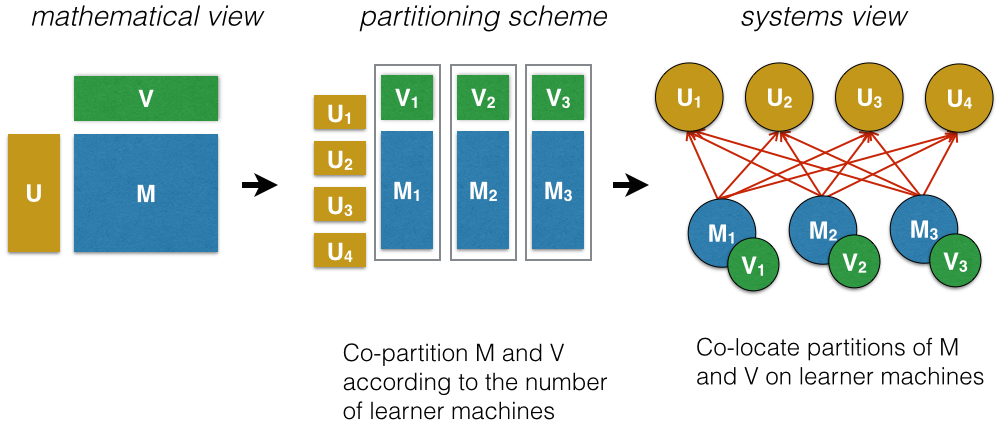}
  \end{center}
  \caption{Partitioning scheme and system architecture.}
  \label{fig:bigpicture}
\end{figure}

This architecture inevitably leads to the second challege, the question of how to handle concurrent, possibly conflicting updates to the factor vectors. When two learner machines fetch, update and write back the same factor vector concurrently, one such update will be overridden.\\
In the special case of matrix factorization, approaches to parallelizing SGD have been  proposed that leverage a carefully chosen  
partitioning of the input data to avoid conflicting updates~\cite{Gemulla2011, Recht2013}. As these approaches require complex data movement  and  synchronization  patterns and are at the same time  hard to adapt to a streaming scenario, we decided for an alternative approach that is simpler to implement in a distributed setting. Instead of taking complex actions to prevent conflicting updates, Factorbird builds upon a recently proposed parallelization scheme for SGD-based learning called Hogwild! \cite{Recht2011}. This work states that parallel SGD can be implemented \textit{without any locking} if most updates only modify small parts of the model. The authors explicitly name latent factor models for matrix factorization as one such case. \\
The special nature of the matrix factorization problem allows us for a further optimization in our system design, which reduces the required network traffic and at the same time greatly lowers the probability for conflicting overwrites. We can reduce the communication cost by 50\% through intelligent partitioning, as follows. If we partition M by either rows or columns over the learning machines, than the updates to either $U$ or $V$ become local, when we co-partition one of the factor matrices ($U$ in the case of partitioning by rows, $V$ in the case of partitioning by columns) with $M$ on the learner machines.\\ 
In the light of minimizing update conflict potential, we decide to co-­locate $V$ on the learner machines and keep $U$ in the parameter machines (c.f. Figure~\ref{fig:bigpicture}). We choose this scheme for the following reasons: In case of the follower graph, the number of updates to a factor vector $u_i$ in $U$ is equal to the  out-­degree of the corresponding vertex $i$, while the number of updates to a factor vector $v_j$ in $V$ is equal to the in-­degree of the corresponding vertex $j$. As the in-­degree distribution of the follower graph has much higher skew than the out-­degree distribution~\cite{Myers2014}, we choose to localize the updates to $V$, which gives us a higher reduction in conflict potential than localizing $U$. Other graphs in twitter have similar skew in the degree distribution.

\section{Implementation}
\label{sec:implementation}

We implement Factorbird using twitter's existing technology stack and infrastructure. We leverage an existing memcached cluster for the parameter machines and implement the learner machines in Scala as a custom finagle application~\cite{Eriksen2013}. We use Hadoop's distributed filesystem (HDFS) as persistent datastore for reading inputs and storing the final results. Prepartitioning and statistics computation of the input is conducted via MapReduce using Scalding jobs. The learner machines are assigned resources via Apache Mesos.\\
The typical execution steps on a learner machine are as follows. First, the machine loads statistics about the input graph and its assigned partition from HDFS, e.g. the number of edges and vertices, the average edge weight and the degree distribution of the graph. These statistics are used for learning as well as efficiently allocating memory for the required datastructures. Next, the learner machine instantiates its local partition of $V$. Subsequently, the model learning phase starts: the learner machine reads the edges of its assigned partition of the input graph in a streaming fashion. For each edge $(i,j)$, it reads the corresponding factor vector $u_i$ from memcached and the factor vector $v_j$ from its local partition of $V$. The reads from memcached are conducted in batches to increase throughput. Next, the learner machine updates the factor vectors using SGD and writes them back\footnote{Note that strictly speaking, we do not run vanilla Hogwild! on $U$, as memcached atomically updates the whole factor vector instead of individually updating the factors.}. This process is repeated until a user-specified number of passes over the input graph has been conducted. Finally, the learner machines persist the learned matrices $U$ and $V$ in a partitioned manner in HDFS.\\
Factorbird makes extensive use of memory-efficient data structures for managing factor vectors, the local partition of $V$ and graph statistics such as the degree per vertex. The main objective of these data structures is to use as little memory as possible as well as to avoid object allocation and full garbage collection invocations in the JVM. A factor vector as well as a partition of a factor matrix are therefore internally represented by byte buffers and large primitive arrays, which are efficiently preallocated. For example, a learning machine determines the size of its local partition of $V$ at startup time by reading the number of vertices assigned to its partition from the graph statistics. The partition of $V$ is internally represented by a huge float array, into which the individual factor vectors (the columns of $V$) are packed. A mapping of vertex ids to offsets in this array is stored and update operations directly write to the underlying array. During the training phase, a learner machine directly reads the training edges from a compressed file in HDFS in a streaming fashion. If more than one pass through the training edges is necessary, the learner machine will on-the-fly create a copy of the HDFS file on local disk and switch to streaming edges from this local file for subsequent passes. Furthermore, some learning approaches require synthetically generated negative examples~\cite{Pan2008} (possibly taken into account with a lower confidence than observed positive examples). We therefore implement on-the-fly generation of such negative examples (with a configurable probability) and mix them into the original positive examples supplied on the learning machines.\\
The central abstraction for implementing the SGD updates of the learning algorithm within Factorbird is the \textit{Learner}~(c.f, Listing~\ref{lst:learner}). It is the main interface for data scientists wanting to try new models and shields the programmer from the complexity of the distributed nature of the underlying learning process. The first important method that a programmer has to implement is \textit{initialize} which Factorbird uses to randomly initialize of the factor vectors in $U$ and $V$ (c.f. line  in Algorithm~\ref{alg:plainsgd}). The method \textit{update} is where the SGD-based update of two factor vectors $u_i$ and $v_j$ is implemented. The system provides the strength $a(i,j)$ of the edge between vertex $i$ and $j$, as well the vertex degrees $n_i$ and $n_j$ and the error weight $w(i,j)$ as additional arguments. A typical implementation of this method conducts the steps from line 4 to 9 in Algorithm~\ref{alg:plainsgd}.

\begin{lstlisting}[caption=Learner abstraction., label=lst:learner, captionpos=b, frame=tb]
trait Learner {
  def initialize(factors: FactorVector): Unit
  def update(u_i: FactorVector, v_j: FactorVector, 
                a_ij: Float, n_i: Int, n_j: Int, w_ij: Float): Float
}
\end{lstlisting}

\section{Assessing Model Quality}
\label{sec:modelquality}

\begin{figure}[h]
  \begin{center}
  \includegraphics[scale=0.4]{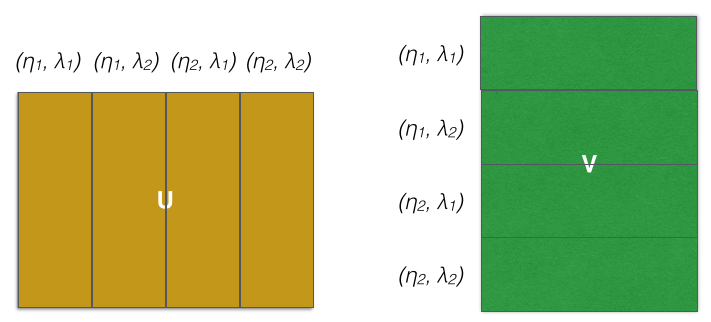}
  \end{center}
  \caption{Packing many models into one for hyperparameter search.}
  \label{fig:packing}
\end{figure}

The next aspect we focus on in Factorbird is the quality of the learned models. Ultimately, the models have to be evaluated using online experiments with real users, but during the development and batch training phase, we concentrate on a simple offline metric: the prediction quality on held-out data, measured by the root mean squared error (RMSE). Unfortunately, this prediction quality is heavily influenced by the choice of hyperparameters for our model, such as $\eta$ (which controls rate of learning), $\lambda$ (which controls regularization), number of factors $k$ as well as the rate of decay of the learning rate in our \textit{Learner} implementation. In order to find a well-working hyperparameter combination, we conduct a grid search in the hyperparameter space. We extend Factorbird to enable hold-out tests at scale. The Scalding job which prepares the input graph randomly splits the edges into training set, validation set and test set. Factorbird then learns a model on the training set, chooses the hyperparameter combination using the prediction quality on the validation set, and finally computes the RMSE on the test set.\\
However, conducting a single training run with Factorbird for each hyperparameter combination to inspect is tedious and
takes a long time. We therefore we describe how to learn many models with different hyperparameters at once to speed up
the hyperparameter search. Given that we aim to inspect $c$ hyperparameter combinations for a factorization of rank $k$,
we pack the $c$ factor vectors into a large $U$ of dimensionality $m \times c*k$ and a large $V$ of dimensionality $c*k
\times n$. We use a specialized learner implementation that is aware of the packing (e.g. it knows that the factors for
the $p$-th model are contained in the $p*k$-th to $p*(k+1)$-th entries of a factor vector) and learns $c$ models at once. Figure~\ref{fig:packing} illustrates how the factor matrices for all possible combinations of two different learning rates $\eta_1$, $\eta_2$ and two different regularization constants $\lambda_1$, $\lambda_2$ would be packed into a large $U$ and $V$.
 
In order to compute the prediction quality on held-out data, Factorbird asks the \textit{Learner} implementation for \textit{Predictor}s which the user has to provide. A \textit{Predictor} predicts the strength of an unobserved edge $(i,j)$ from the corresponding factor vectors $u_i$ and $v_j$ (c.f., Listing~\ref{lst:predictor}). 

\begin{lstlisting}[caption=Predictor abstraction., label=lst:predictor, captionpos=b, frame=tb]
trait Predictor {
  def predict(u_i: FactorVector, v_j: FactorVector): Float
}
\end{lstlisting}

Optionally, the user can provide a \textit{LossEstimator}, which estimates the current value of the loss function using samples of edges and factors~(c.f., Listing~\ref{lst:loss}). During training, Factorbird continually invokes the loss estimator and make the estimates inspectable via an external dashboard.

 \begin{lstlisting}[caption=Abstraction for loss estimation., label=lst:loss, captionpos=b, frame=tb]
trait LossEstimator {
  def estimateRegularizationComponent(
    numRowsOfU: Int, sampleOfU: Iterator[FactorVector],
    numColumnsOfV: Int, sampleOfV: Iterator[FactorVector]): Double
  
  def estimateErrorComponent(numEdges: Long,
    sampleOfEdges: Iterator[Edge], partitionOfU: FactorMatrix,
    partitionOfV: FactorMatrix): Double
}
\end{lstlisting}

\section{Experiments}
\label{sec:experiments}

We run experiments on various subsets of `RealGraph', a graph that models various interactions between twitter users~\cite{Kamath2014}. The learner machines for our experiments are provisioned by Apache Mesos. In all our experiments, we factorize the binarized adjacency matrix of the graph subset. That means the, transformation function $a(i,j)$ returns $1$ if user $i$ interacted with user $j$ and $0$ otherwise (in the case of a synthetic negative example). We equally weight all prediction errors ($w(i,j) = 1$). In this work, we present only preliminary experiments, aimed at validating the correctness of our system and showing its capacity to handle twitter-scale graphs. There is still a large potential for improvements in accuracy and performance that we will tackle in future work. We run a first set of experiments on a small sample of the RealGraph, consisting of 100 million interactions between 440 thousand popular users. Additionally, we make Factorbird generate 500 million synthetic negative edges. 

\begin{figure}[h]
	\centering
	\begin{minipage}[t]{0.48\textwidth}
		\vspace{0pt}
		\centering
    \includegraphics[scale=0.35]{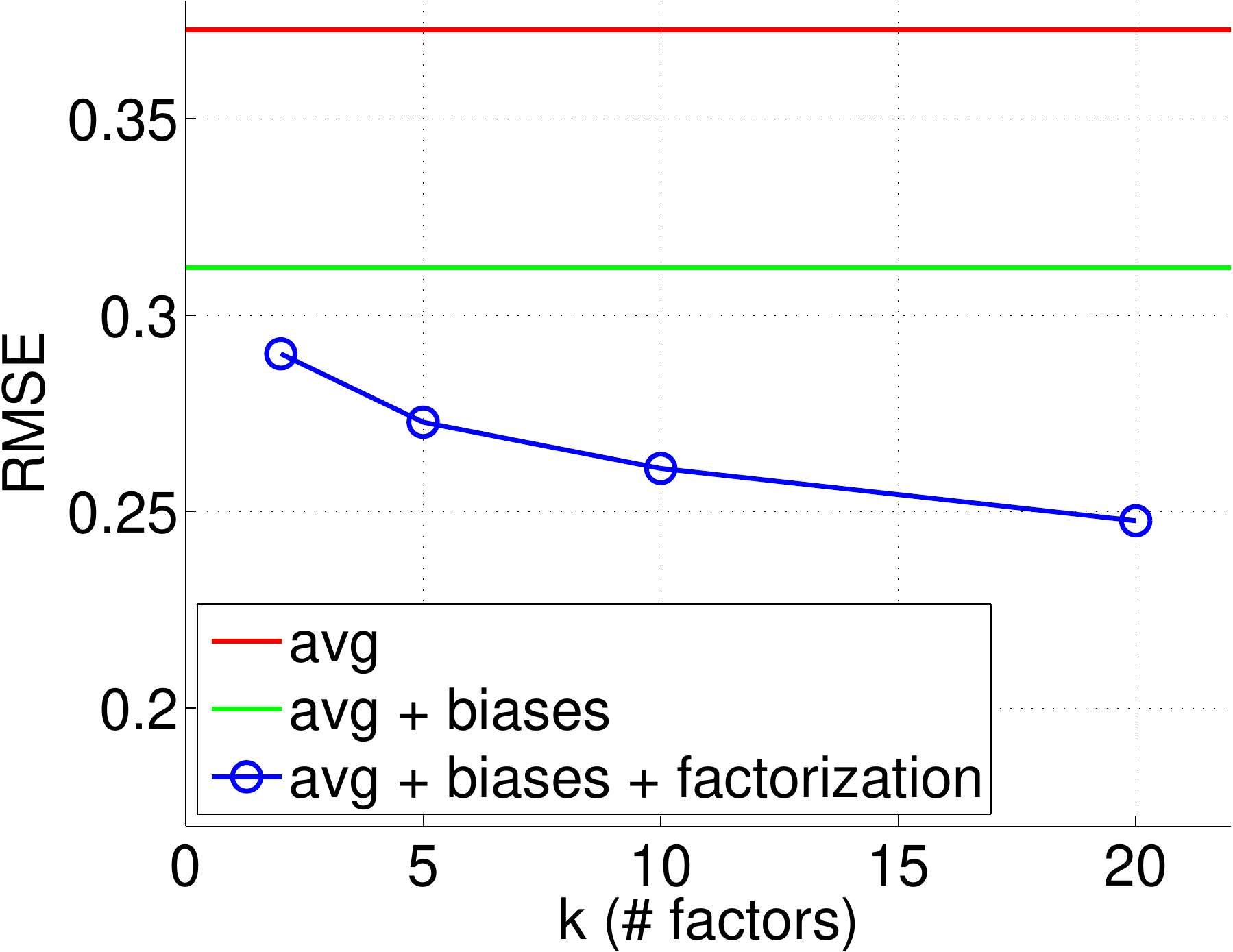}
    \caption{Prediction quality on held-out data with increasing model complexity.}
    \label{fig:validation}
	\end{minipage}
	\hfill
	\begin{minipage}[t]{0.48\textwidth}
		\vspace{0pt}
		\centering
  \includegraphics[scale=0.35]{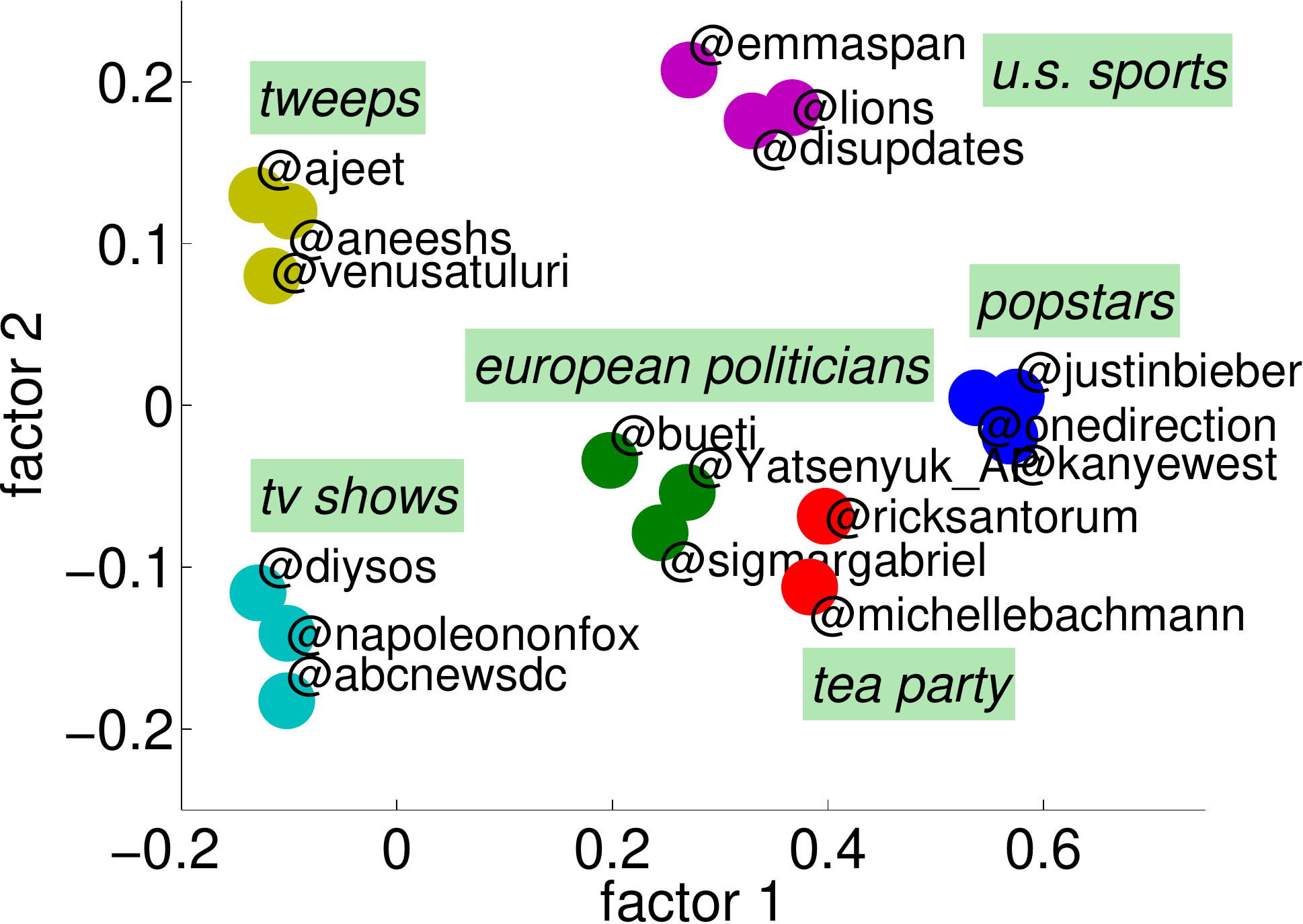}
    \caption{Plot of a selection of twitter users as positioned by a factorization with $k = 2$ of a sample of RealGraph.}
    \label{fig:plot}
	\end{minipage}
%\vspace{-12pt}
\end{figure}

\textbf{Benefits of increasing model complexity.} In the first experiment, we show the positive effects on prediction quality of the individual parts of our chosen model. We randomly split the dataset into 80\% training set, 10\% validation set and 10\% test set. We train models with increasing complexity and measure their prediction quality in terms of RMSE on the 10\% held-out data in the test set (c.f.~Figure~\ref{fig:validation}). We start with a baseline that only uses the global bias (the average edge strength), followed by a more complex model that uses the global bias as well as vertex-specific bias terms. Finally, we train biased factorization models with $k \in \{2, 5, 10, 20\}$. We choose the hyperparameters using the validation set. The outcome confirms that an increase in the complexity of our model results in an increase in prediction quality. The global bias baseline provides an RMSE of $0.3727$, adding the vertex-specific bias terms reduces the error to $0.3121$ and incorporating the factors gives additional improvements, reducing the error to $0.2477$ for $k = 20$.

\textbf{Visual inspection.} Next, we plot a selection of factor vectors from the $V$ matrix of a biased factorization with $k = 2$. Two users $i$ and $j$ will be close in the resulting low-dimensional space if their follower vectors $m_{\bullet i}$ and $m_{\bullet j}$ are roughly linear combinations of each other. Due to homophily, we expect twitter users sharing a common interest or characteristic to have a large overlap in followers and therefore to be much closer in the resulting space. Figure~\ref{fig:plot} indicates that factorizations produced by Factorbird have this property. We see several clusters of twitter accounts of similar type, e.g. a cluster of european politicians, containing Sigmar Gabriel ({\tt @sigmargabriel}) and Reinhard Buetikofer ({\tt @bueti}) from Germany as well as Arseniy Yatsenyuk ({\tt @Yatsenuk\_AP}) from Ukraine. Another cluster consists of popstars such as Kanye West ({\tt @kanyewest}), Justin Bieber ({\tt @justinbieber}) and the boy-band \textit{One Direction} ({\tt @onedirection}). A third one related to U.S. sports contains Emma Span ({\tt @emmaspan}), an editor for baseball at Sports Illustrated, Detroit's football team ({\tt @lions}) and an account of the Daytona International Speedway ({\tt @disupdates}).

\textbf{Scale-out.} For the scale-out experiments we build a matrix based on a large subset of Twitter's
RealGraph~\cite{Kamath2014} consisting of more than 38.5 billion non-zeros (half of which are synthetic negative
examples). The dimensions of the matrix are 229 million $\times$ 195 million. To the best of our knowledge, this is the largest dataset on which collaborative filtering experiments have been published so far. Often, the Netflix prize dataset~\cite{Bennett2007} is used for scale-out tests \cite{Low2012}. Our dataset is more than two orders of magnitude larger, having approximately 470 times more rows, 11,000 times more columns and 385 times more datapoints. We run Factorbird using 50 instances for the learner machines, provisioned in a Mesos cluster, with 16 cores and 16 gb of memory each. We leverage a shared memcached cluster for our parameter machines, with a guaranteed quota of 5 million commands per second. We run two passes of hyperparameter search on 80\% of the data for 16 different learners with $k$ = 5. The rank of the factor matrices $U$ and $V$ is $(5+1) * 16 = 96$, which means we train a model with approximately $(229M + 195M) \; * \; 96 \approx 40B$ parameters. A single SGD pass through the training set finishes in about 2.5 hours~\footnote{Except for a few stragglers, which Mesos provisioned on machines with low network bandwidth (Unfortunately, Mesos cannot guarantee a defined network bandwidth at the moment)}. In this experiment, Factorbird issues more than 4.2 million commands per second to memcached on average and updates around 400M parameters per second.

\section{Related Work}

SGD- and ALS-based matrix factorization techniques for recommender systems have been extensively studied in the context of the Netflix Prize \cite{Koren2009, Zhou2008}. These techniques have been extended to work on implicit feedback data \cite{Hu2008} and to optimize metrics different from RMSE \cite{Rendle2009, Weimer2007}.\\
A large body of work has been conducted with respect to parallelizing and distributing matrix factorization. This includes work on the scalability of the algorithms itself, e.g. by introducing biased sampling for SGD to avoid conflicting updates  during concurrent execution \cite{Gemulla2011, Teflioudi2012, Recht2013} or by proving convergence under a minor amount of update conflicts \cite{Recht2011}. Furthermore, distributed implementations have been proposed and evaluated on MapReduce-based \cite{Schelter2013}, graph-parallel \cite{Low2012, Xin2014} and specialized systems \cite{Yun2014}.

\section{Future Work}

In future work, we aim to extend Factorbird to a streaming scenario. We plan to bootstrap Factorbird with a factorization that was trained offline and update this factorization model online from a stream of incoming real-time interactions (e.g., user follows). Furthermore, we would like to experiment with factorizing multiple matrices at once, in order to incorporate different types of interactions in a single factorization model. Another technical issue to work on in the future is fault tolerance. A possible approach to recovery in case of failures could be to restart learner and parameter machines from asynchronously written checkpoints of the partitions of $U$ and $V$ which they hold~\cite{SchelterCIKM2013}. Moreover, we plan to investigate ways to reduce the amount of network traffic caused by Factorbird, e.g. by compression, factor vector caching or via a biased sampling of the edges to allow us to use retrieved factor vectors for more than a single update. We will potentially replace memcached with a custom application to be able to achieve higher throughput and conduct true Hogwild-style updates on the parameter machines. Moreover, this would allow us to run aggregations on the parameter machines. Additionally, we would like to implement dynamic load adaption in Factorbird to mitigate the negative effects of stragglers on the overall runtime. We aim to factorize the whole twitter interaction graph with Factorbird in upcoming work.

\newpage
\small{

}


\begin{thebibliography}{10}
\providecommand{\url}[1]{#1}
\csname url@samestyle\endcsname
\providecommand{\newblock}{\relax}
\providecommand{\bibinfo}[2]{#2}
\providecommand{\BIBentrySTDinterwordspacing}{\spaceskip=0pt\relax}
\providecommand{\BIBentryALTinterwordstretchfactor}{4}
\providecommand{\BIBentryALTinterwordspacing}{\spaceskip=\fontdimen2\font plus
\BIBentryALTinterwordstretchfactor\fontdimen3\font minus
  \fontdimen4\font\relax}
\providecommand{\BIBforeignlanguage}[2]{{%
\expandafter\ifx\csname l@#1\endcsname\relax
\typeout{** WARNING: IEEEtran.bst: No hyphenation pattern has been}%
\typeout{** loaded for the language `#1'. Using the pattern for}%
\typeout{** the default language instead.}%
\else
\language=\csname l@#1\endcsname
\fi
#2}}
\providecommand{\BIBdecl}{\relax}
\BIBdecl

\bibitem{Amatriain2012}
Amatriain,~X. (2012) Building industrial-scale real-world recommender systems. In {\it ACM Conference on Recommender Systems},  \url{http://goo.gl/cqKAi}

\bibitem{Koren2009}
Koren, Y., Bell, R. \& Volinsky C. (2009) Matrix factorization techniques for recommender systems. In {\it Computer 42.8}, pp. 30--37.

\bibitem{Bennett2007}
Bennett,~J. \& Lanning,~S. (2007) The Netflix Prize. In {\it KDD Cup and Workshop}, p. 35.

\bibitem{Schelter2013}
Schelter, S., Boden, C., Schenck, M., Alexandrov, A. \& Markl, V. (2013). Distributed matrix factorization with mapreduce using a series of broadcast-joins. In {\it ACM Conference on Recommender Systems}, pp. 281-284.

\bibitem{Teflioudi2012}
Teflioudi, C., Makari, F. \& Gemulla, R. (2012) Distributed Matrix Completion. In {\it IEEE International Conference on Data Mining}, pp. 655-664.

\bibitem{Das2007}
Das, A. S., Datar, M., Garg, A. \& Rajaram, S. (2007). Google news personalization: scalable online collaborative filtering. In {\it International Conference on the World Wide Web}, pp. 271-280.

\bibitem{Gemulla2011}
Gemulla, R., Nijkamp, E., Haas, P. J. \& Sismanis, Y. (2011). Large-scale matrix factorization with distributed stochastic gradient descent. In {\it ACM Conference on Knowledge Discovery and Data mining}, pp. 69-77. top of the rock new york

\bibitem{Hu2008}
Hu, Y., Koren, Y. \& Volinsky, C. (2008). Collaborative filtering for implicit feedback datasets. In {\it IEEE International Conference on Data Mining}, pp. 263-272.

\bibitem{Zhou2008}
Zhou, Y., Wilkinson, D., Schreiber, R. \& Pan, R. (2008). Large-scale parallel collaborative filtering for the netflix prize. In {\it Algorithmic Aspects in Information and Management}, pp. 337-348.

\bibitem{Rendle2009}
Rendle, S., Freudenthaler, C., Gantner, Z. \& Schmidt-Thieme, L. (2009). BPR: Bayesian personalized ranking from implicit feedback. In {\it Conference on Uncertainty in Artificial Intelligence}, pp. 452-461.

\bibitem{Low2012}
Low, Y., Bickson, D., Gonzalez, J., Guestrin, C., Kyrola, A. \& Hellerstein, J. M. (2012). Distributed GraphLab: a framework for machine learning and data mining in the cloud. In {\it Proceedings of the VLDB Endowment}, 5(8), pp. 716-727.

\bibitem{Xin2014}
Xin, R. S., Crankshaw, D., Dave, A., Gonzalez, J. E., Franklin, M. J. \& Stoica, I. (2014). GraphX: Unifying Data-Parallel and Graph-Parallel Analytics. {\it arXiv preprint arXiv:1402.2394}.

\bibitem{Yun2014}
Yun, H., Yu, H. F., Hsieh, C. J., Vishwanathan, S. V. N. \& Dhillon, I. (2013). NOMAD: Non-locking, stOchastic Multi-machine algorithm for Asynchronous and Decentralized matrix completion. {\it Proceedings of the VLDB Endowment}, 7(11).

\bibitem{Recht2011}
Recht, B., Re, C., Wright, S. \& Niu, F. (2011). Hogwild!: A lock-free approach to parallelizing stochastic gradient descent. In {\it Advances in Neural Information Processing Systems}, pp. 693-701.

\bibitem{Recht2013}
Recht, B. \& Re, C. (2013). Parallel stochastic gradient algorithms for large-scale matrix completion. In {\it Mathematical Programming Computation}, 5(2), pp. 201-226.

\bibitem{Weimer2007}
Weimer, M., Karatzoglou, A., Le, Q. V. \& Smola, A. (2007). Maximum Margin Matrix Factorization for Collaborative Ranking. In {\it Advances in Neural Information Processing Systems}.

\bibitem{Li2014}
Li, M., Zhou, L., Yang, Z., Li, A., Xia, F., Andersen, D. G. \& Smola, A. (2014) Parameter Server for Distributed Machine Learning.

\bibitem{Eriksen2013}
Eriksen, M. (2013). Your server as a function. In {\it Workshop on Programming Languages and Operating Systems},  p. 5.

\bibitem{SchelterCIKM2013}
Schelter, S., Ewen, S., Tzoumas, K. \& Markl, V. (2013). ``All Roads Lead to Rome:'' Optimistic Recovery for Distributed Iterative Data Processing. In {\it ACM International Conference on Information \& Knowledge Management}, pp. 1919-1928.

\bibitem{Kamath2014}
Kamath, K.,  Sharma, A., Wang, D. \& Yin, Z. (2014) RealGraph: User Interaction Prediction at Twitter. In {\it User Engagement Optimization Workshop @ KDD}

\bibitem{Myers2014}
Myers, S. A., Sharma, A., Gupta, P. \& Lin, J. (2014). Information network or social network?: the structure of the twitter follow graph. In {\it International Conference on the World Wide Web}, pp. 493-498.  
 
\bibitem{Hastie2014}
Hastie, T.,  Mazumder, R., Lee, J. D. Lee \& Zadeh, R. (2014). Matrix Completion and Low-Rank SVD via Fast Alternating Least Squares. In {\it arXiv}
 
\bibitem{Boyd2014}
Udell, M., Horn, C., Zadeh, R. \& Boyd, S. (2014). Generalized Low Rank Models. In {\it arXiv}
  
\bibitem{Pan2008}
Pan, R., Zhou, Y., Cao, B., Liu, N. N., Lukose, R., Scholz, M. \& Yang, Q. (2008). One-class collaborative filtering. In {\it IEEE International Conference on Data Mining}, pp. 502-511.
  
\end{thebibliography}
\end{document}